\documentclass[11pt,a4paper]{article}
\usepackage[utf8]{inputenc}
\usepackage[T1]{fontenc}
\usepackage{newtxtext,newtxmath}
\usepackage[margin=1in]{geometry}
\usepackage{microtype,setspace,amsmath,booktabs,array,graphicx,float,hyperref,caption,enumitem}
\usepackage[numbers,sort&compress]{natbib}
\hypersetup{colorlinks=true,linkcolor=black,citecolor=black,urlcolor=black,pdftitle={Layer-Wise Gate-Controlled Prompt Truncation in a Multimodal Chest X-Ray Classifier},pdfauthor={Jingtao Lei; Hongji Li; Dexiang Shu}}
\newcommand{\R}{\mathbb{R}}
\begin{document}
\begin{center}
\vspace*{1.1em}
{\LARGE\bfseries Layer-Wise Gate-Controlled Prompt Truncation in a Multimodal Chest X-Ray Classifier\par}
\vspace{1.5em}
{\large Jingtao Lei\textsuperscript{1}, Hongji Li\textsuperscript{2}, Dexiang Shu\textsuperscript{3}\par}
\vspace{0.8em}
{\small
\textsuperscript{1}Central South University, Changsha 410083, China; 2617549@dundee.ac.uk\\
\textsuperscript{2}Hong Kong Baptist University, Hong Kong, China; 23263970@life.hkbu.edu.hk\\
\textsuperscript{3}University of Minnesota, Minneapolis, MN 55414, USA; shu00038@umn.edu}
\end{center}
\vspace{0.6em}
\begin{abstract}
Mixture of Prompt Experts (MoPE) adapts multimodal transformers through input-dependent prompt composition, while retaining a fixed prompt length. We investigate a layer-wise gating extension in a binary chest X-ray classification pilot study. The controller predicts a retention ratio for each sample, averages these ratios within a mini-batch, and uses the resulting integer length to truncate the static and mixed visual prompts. Retained mixed prompts are also scaled by the individual ratios. In one recorded run per configuration, the gated model reached a best validation accuracy of 0.8996, compared with 0.8969 for the fixed-length baseline; the corresponding final values were 0.8963 and 0.8802. The exported gate statistics imply a retained length of one at all recorded training points, relative to a configured maximum of six. This reduces the complete visual sequence from 210 to 200 tokens, but no direct runtime measurements establish an acceleration benefit. Report-derived labels, report text as input, sequential data partitioning, and the absence of repeated controlled experiments limit interpretation. The findings document prompt shortening under the configured gate penalty; they do not establish sample-specific length allocation, superiority over fixed short prompts, or clinical utility. Code is available at: \url{https://github.com/jingtaolei/mope-dynamic-prompt-truncation}.
\end{abstract}
\noindent\textbf{Keywords:} multimodal learning; visual prompt tuning; mixture of prompt experts; prompt truncation; parameter-efficient adaptation; chest X-ray classification

\section{Introduction}
Transformer architectures underpin many pretrained language and vision models \cite{vaswani,bert,vit}. Adapting these models to a new task can require substantial task-specific storage and optimization. Continuous prompt tuning addresses this problem by learning additional token representations while retaining pretrained backbone weights \cite{lester,jia,prefix,ptuning}. Low-rank adaptation provides an alternative through trainable low-rank weight updates \cite{lora}. These methods motivate the study of adaptation capacity as an explicit design choice rather than an unrestricted consequence of full fine-tuning.

Multimodal adaptation must also coordinate information across input modalities. Mixture of Prompt Experts (MoPE) represents adaptation prompts as combinations of expert prompt banks, with routing weights conditioned on multimodal features \cite{mope}. This mechanism varies prompt composition across samples. However, the number of prompt tokens supplied to a transformer layer remains fixed. Prompt composition and prompt length therefore represent different dimensions of adaptation.

We investigate a gate-controlled extension that changes the retained visual prompt length at each layer. A two-layer feed-forward network predicts a ratio from the current visual class token and a mapped report representation. The implementation averages these ratios over the mini-batch before discretization, so each batch shares one retained length at a given layer. The sample-specific ratios additionally scale the mixed prompts. The controller therefore combines batch-dependent length selection with sample-specific amplitude modulation.

We examine this mechanism in a derived normal-versus-abnormal classification task using chest radiographs and reports. The study focuses on prompt behavior in this specific multimodal optimization setting. The construction of the report-derived target and its overlap with the text input are detailed in Section~\ref{sec:data}.

The study contributes a precise account of the implemented gate and its gradient paths, a reanalysis of the recorded comparison with a fixed-length model, and explicit accounting of prompt tokens and controller parameters. The observed gate statistics are consistent with minimum-length collapse rather than demonstrated allocation according to input difficulty. We therefore examine whether the evidence supports prompt shortening, and identify the controls required to establish an advantage over a short fixed prompt.

\section{Related Work}
\subsection{Parameter-Efficient Adaptation}
Language prompt tuning learns continuous input representations, while prefix tuning introduces learned representations into transformer computation \cite{lester,prefix}. Deep prompting extends this idea across layers and tasks \cite{ptuning}. Visual prompt tuning applies additional trainable tokens to a frozen vision transformer \cite{jia}. These approaches separate the adaptation parameters from the pretrained backbone. Prompt length is a further design variable that controls how many additional representations enter each layer.

\subsection{Prompt Mixtures and Multimodal Fusion}
MoPE uses multimodal routing to construct input-dependent combinations of prompt experts \cite{mope}. Its expert-mixture mechanism supplies the foundation for the present classifier. The modification studied here concerns truncation and amplitude scaling of its visual prompts; the multimodal expert router is retained. An auxiliary importance loss encourages aggregate expert-use balance.

\subsection{Adaptive Transformer Computation}
DynamicViT sparsifies visual content tokens, EViT reorganizes inattentive tokens, and AdaViT selects computation through input-dependent policies \cite{dynamicvit,evit,adavit}. The present mechanism instead shortens inserted prompt prefixes while retaining every image-patch token. It therefore targets the adaptation sequence rather than the spatial resolution of visual content tokens. Section~\ref{sec:accounting} quantifies how this distinction affects the fraction of the complete sequence that can be removed.

\subsection{Medical Image--Text Representation Learning}
ConVIRT and GLoRIA learn medical image representations from paired radiographs and reports through contrastive objectives \cite{convirt,gloria}. MedKLIP incorporates medical knowledge into language--image pretraining, and PTUnifier uses soft prompts to support different modality combinations \cite{medklip,ptunifier}. These studies motivate multimodal representation learning, but their evaluation settings differ from a classifier that receives the same report text used to derive its target. For the latter setting, unimodal controls and independently defined outcomes are essential to interpreting multimodal performance.

\section{Materials and Methods}
\subsection{Data Source and Derived Classification Task}
\label{sec:data}
The pilot study uses a Kaggle-organized version of the Indiana University chest X-ray collection, derived from the public Open-I radiology resource \cite{iu}. The loader reads \texttt{indiana\_reports.csv} and \texttt{indiana\_projections.csv}, joins the tables on the study identifier \texttt{uid}, and treats each resulting projection--report pair as a classification sample. Missing findings and impression fields are replaced with empty strings before concatenation to form the text input.

The implemented target rule assigns label 0 when the lowercased MeSH field contains the substring \texttt{normal}, or the lowercased impression contains \texttt{no acute} or \texttt{unremarkable}; otherwise it assigns label 1. We refer to these categories as rule-derived normal and abnormal labels. They are not independently adjudicated clinical endpoints. In particular, substring matching does not distinguish \texttt{normal} from \texttt{abnormal}, and the phrase \texttt{no acute} does not exclude chronic abnormalities. These properties introduce label noise in addition to the overlap between the target-generation fields and model input.

After merging, the first $\lfloor0.8n\rfloor$ of the $n$ rows are assigned to training and the remaining rows to validation. This partition is neither randomized nor stratified and does not enforce study- or patient-level separation. Training batches are subsequently shuffled, which does not change the partition or establish group independence. The same non-training subset is also returned by the nominal test loader; its scores are therefore reported throughout as validation results. Exact post-merge cohort counts, class counts, and overlap statistics are unavailable in the retained scalar records.

Images are read as RGB PNG files from the configured image directory. Training and validation use the same deterministic preprocessing: resize the shorter side to 256 pixels, center-crop to $224\times224$, convert to a tensor, and normalize the three channels using means $(0.46777044,0.44531429,0.40661017)$ and standard deviations $(0.12221994,0.12145835,0.14380469)$. These are the constants in the supplied loader; their derivation for this cohort is not documented. The training loader drops an incomplete final batch, whereas validation retains all rows. If image reading fails, the loader supplies a black RGB image before preprocessing. The records do not establish how often this fallback was activated.

\subsection{Base Architecture and Prompt Composition}
The visual encoder is a pretrained ViT-B/16 with 12 transformer layers and embedding dimension $D=768$ \cite{vit}. The text branch uses \texttt{bert-base-uncased} \cite{bert}; reports are tokenized with padding and truncation, and four prompt tokens are inserted at each text layer. The supplied launch command leaves \texttt{train\_instructor} disabled. Consequently, the entire text branch, including its uniformly initialized prompt parameters, is frozen and evaluated without gradient tracking. These text prompts are not learned in this configuration. Adaptation updates the visual prompt banks, internal router projections, instruction mapper, gate, and visual classification head while retaining the pretrained visual backbone weights. The configured maximum visual prompt length is $P=6$, and the number of experts is $K=4$.

Let $E_{\ell,k}\in\R^{P\times D}$ denote expert $k$ at visual layer $\ell$, and let $A_\ell\in\R^{P\times D}$ denote the static prompt bank. For sample $i$, the router forms the mixed prompt
\begin{equation}
 M_{i,\ell}=\sum_{k=1}^{K}s_{i,\ell,k}E_{\ell,k},
 \qquad \sum_{k=1}^{K}s_{i,\ell,k}=1.
\label{eq:mixture}
\end{equation}
The internal router projects mean-pooled visual patch features and the BERT-derived text feature into dimensions two and eight, respectively. Their concatenation is multiplied by a frozen expert-key matrix, and a softmax with temperature 0.1 produces the expert weights. The implementation adds Gaussian routing noise with scale $1/K^2$ after temperature scaling. This noise is not conditioned on training mode in the archived code and therefore also introduces stochasticity during evaluation.

The text feature $t_i\in\R^{768}$ is separately mapped to an instruction vector $z_i\in\R^{D}$ through a $768\rightarrow640\rightarrow768$ projection with Gaussian error linear unit (GELU) activations. This vector conditions the gate and is also inserted as an instruction token. At the first visual layer, the class token has not yet attended to image patches; the first gate should therefore not be interpreted as measuring image difficulty from a contextualized visual representation.

\subsection{Gate Prediction and Batch-Dependent Truncation}
For the visual class token $c_{i,\ell}$ and mapped instruction $z_i$, the gate computes
\begin{equation}
 r_{i,\ell}=\sigma\!\left(W_{\ell,2}\operatorname{ReLU}
 \left(W_{\ell,1}[c_{i,\ell};z_i]+b_{\ell,1}\right)+b_{\ell,2}\right),
\label{eq:gate}
\end{equation}
where $W_{\ell,1}\in\R^{192\times1536}$, $W_{\ell,2}\in\R^{1\times192}$, and $\sigma$ denotes the sigmoid function. For mini-batch $\mathcal{B}$, the retained length is
\begin{equation}
 \bar r_{\mathcal{B},\ell}=\frac{1}{|\mathcal{B}|}\sum_{i\in\mathcal{B}}r_{i,\ell},
 \qquad L_{\mathcal{B},\ell}=\min\!\left(P,\max\!\left(1,
 \left\lceil P\bar r_{\mathcal{B},\ell}\right\rceil\right)\right).
\label{eq:length}
\end{equation}
Both the mixed prompt and static prompt are sliced to this common prefix length. The retained mixed prompt is additionally scaled by the sample-specific ratio:
\begin{equation}
 \widetilde M_{i,\ell}=r_{i,\ell}M_{i,\ell}[1:L_{\mathcal{B},\ell},:],
 \qquad \widetilde A_{\mathcal{B},\ell}=A_\ell[1:L_{\mathcal{B},\ell},:].
\label{eq:scaled}
\end{equation}
The inserted token sequence is the concatenation of $\widetilde A$, $\widetilde M$, and the instruction token, followed by GELU and dropout with probability 0.1. These tokens are placed between the class token and image-patch tokens. After each transformer block, the prompt outputs are discarded and the updated class and patch tokens proceed to the next layer.

\begin{figure}[!htbp]
\centering
\includegraphics[width=\linewidth]{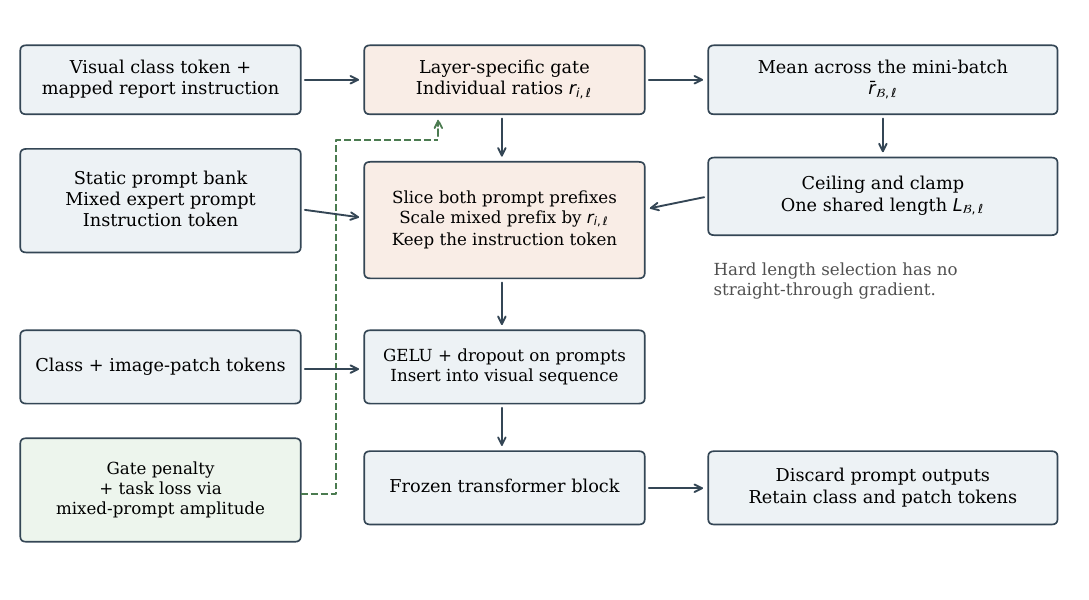}
\caption{Implemented gate-controlled prompt computation at a visual transformer layer. The gate produces individual retention ratios, but their mini-batch mean determines one shared prefix length. Static and mixed prompts are both truncated; only the mixed prompt is multiplied by the individual ratios. Class and image-patch tokens are retained after the block. Dashed arrows identify gradient paths to the gate, rather than gradients through the discrete length decision.}
\label{fig:architecture}
\end{figure}

Figure~\ref{fig:architecture} shows the actual computation. The integer conversion and slicing operation do not provide a derivative with respect to the selected length. Gradients reach the gate through the multiplicative factor in Equation~\eqref{eq:scaled} and through the gate regularizer described below. This is a differentiable amplitude path combined with hard truncation; the archived implementation does not use a straight-through estimator for the discrete length. It also does not partition a mini-batch into groups with different lengths.

\noindent\fbox{\begin{minipage}{\dimexpr\linewidth-2\fboxsep-2\fboxrule\relax}
\small
\textbf{Algorithm 1. Layer-wise gate-controlled prompt truncation.}\\
\textbf{Input:} A mini-batch of images, reports, and training labels.\\
\textbf{Output:} Class logits and, during training, the total loss.
\begin{enumerate}[leftmargin=1.5em,itemsep=0.1em,topsep=0.3em]
\item Encode reports with the frozen text branch; initialize visual tokens and map the text feature to the instruction token.
\item For each visual layer, compute expert weights and the mixed prompt using Equation~\eqref{eq:mixture}; then perform steps 3--5.
\item Predict individual gate ratios; average across the batch and discretize using Equation~\eqref{eq:length}.
\item Slice the static and mixed prompt prefixes to the shared length; scale the mixed prefix by each sample's ratio.
\item Insert the prompts and instruction token, execute the transformer block, and remove the prompt outputs.
\item After the last layer, normalize and classify the class token. Aggregate the layer-wise auxiliary terms and compute the training loss using Equation~\eqref{eq:loss}.
\end{enumerate}
\end{minipage}}

\subsection{Training Objective}
The optimization objective combines classification, expert-balance, and gate penalties:
\begin{equation}
 \mathcal{L}=\mathcal{L}_{\mathrm{cls}}+
 \lambda_{\mathrm{imp}}\mathcal{L}_{\mathrm{imp}}+
 \lambda_{\mathrm{gate}}\mathcal{L}_{\mathrm{gate}},
 \qquad
 \mathcal{L}_{\mathrm{gate}}=\frac{1}{H}\sum_{\ell=1}^{H}\bar r_{\mathcal{B},\ell},
 \quad H=12.
\label{eq:loss}
\end{equation}
Classification uses negative log-likelihood after log-softmax. For the internal visual router, the importance term is the mean across layers of the squared coefficient of variation of batch-summed expert weights. The implementation uses a stabilizing denominator offset of $10^{-6}$ and sets layer contributions to zero when they do not exceed 0.1. The supplied gated configuration sets $\lambda_{\mathrm{imp}}=0.01$ and $\lambda_{\mathrm{gate}}=0.5$.

The gate penalty is the mean retention ratio, rather than the exact number of executed tokens. A ratio approaching zero still yields one retained token because of the minimum-length constraint. At the same time, it attenuates the mixed prompt toward zero through Equation~\eqref{eq:scaled}. The intervention therefore changes both prompt length and prompt amplitude. Their separate effects cannot be identified without dedicated ablations.

\subsection{Token and Parameter Accounting}
\label{sec:accounting}
For $224\times224$ images with $16\times16$ patches, the visual sequence contains 196 patch tokens and one class token. With static prompts and one instruction token enabled, its length at a layer is
\begin{equation}
 N_{\mathcal{B},\ell}=197+2L_{\mathcal{B},\ell}+1.
\label{eq:tokens}
\end{equation}
Thus $L=6$ and $L=1$ correspond to 210 and 200 total visual tokens, respectively. Table~\ref{tab:budget} distinguishes reduction of a prompt component from reduction of the complete visual sequence. These are arithmetic consequences of the architecture, not hardware measurements.

\begin{table}[H]
\centering\small
\caption{Analytical token counts per visual layer. Relative reductions use the fixed-length configuration as the denominator.}
\label{tab:budget}
\begin{tabular}{lrrr}
\toprule
Token component & $L=6$ & $L=1$ & Reduction (\%)\\
\midrule
Mixed prompt & 6 & 1 & 83.33\\
Static prompt & 6 & 1 & 83.33\\
All inserted tokens, including instruction & 13 & 3 & 76.92\\
Complete visual sequence & 210 & 200 & 4.76\\
\bottomrule
\end{tabular}
\end{table}

The static and expert banks contain $H(K+1)PD=276{,}480$ parameters. Each gate contains $1536\times192+192+192+1=295{,}297$ parameters; twelve gates add $3{,}543{,}564$ parameters. The internal router projections contain 92,280 parameters, the instruction mapper contains 984,448, and the two-class visual head contains 1,538. With the text branch and visual backbone frozen, these components give 1,354,746 trainable parameters for the non-gated architecture and 4,898,310 after adding the gate. These are analytical counts for the supplied configuration, not checkpoint-derived measurements of the recorded runs. Physical slicing reduces the executed prefix but does not shrink the allocated parameter banks.

\section{Experimental Setup}
\subsection{Run Configuration and Available Records}
The comparison uses one fixed-length baseline run and one gated run. Each has 31 exported validation-accuracy records and 149 exported records for the principal training scalars. The training records occur at global steps 49, 99, and subsequent intervals of 50 steps through 7449; the final validation record is at step 7460. A separate sanity-check run is excluded from the comparison because it is not identified as the fixed-prompt baseline and its configuration is not a matched experimental control.

Table~\ref{tab:configuration} summarizes the supplied gated launch configuration and the corresponding code settings. The archive includes a non-gated visual implementation and the fixed-baseline scalar exports, but no immutable checkpoint-to-source manifest for both runs. The comparison consequently remains a descriptive analysis of the recorded experiments, rather than a fully controlled retraining study.

\begin{table}[!htbp]
\centering\small
\caption{Documented gated-run configuration. Values reflect the launch file and archived code; an independently captured run environment is not available.}
\label{tab:configuration}
\begin{tabular}{p{0.43\linewidth}p{0.49\linewidth}}
\toprule
Setting & Value\\
\midrule
Visual encoder & ViT-B/16; 12 layers; dimension 768\\
Text encoder & BERT-base-uncased; entire branch frozen\\
Visual / text prompt length & 6 maximum / 4\\
Prompt experts & 4\\
Gate hidden dimension & 192\\
Batch size / random seed & 8 / 42\\
Training duration & 10 epochs\\
Optimizer & AdamW\\
Initial learning rate, active modules & $4\times10^{-4}$\\
Configured text learning rate & $5\times10^{-4}$; inactive for frozen parameters\\
Weight decay & $10^{-3}$ for both parameter groups\\
Learning-rate schedule & Multiply by 0.4 at epochs 3 and 6\\
Importance / gate loss weight & 0.01 / 0.5\\
Numerical precision / gradient clipping & 16-bit mixed precision / norm 1.0\\
Validation schedule & At approximately one-third-epoch intervals\\
\bottomrule
\end{tabular}
\end{table}

The dependency file lists PyTorch 2.8.0, torchvision 0.23.0, PyTorch Lightning 2.0.0, timm 0.9.16, and Transformers 4.34.0. These versions describe the archived dependency specification; the exact installed versions and GPU used for each recorded experiment have not been independently established. The visual learning-rate exports agree between the two main runs and decrease from $4.0\times10^{-4}$ to $6.4\times10^{-5}$.

\subsection{Metrics and Reanalysis}
Validation accuracy is accumulated over the non-training loader. ``Best validation accuracy'' denotes the maximum recorded value, and ``final validation accuracy'' denotes the last exported value. These summaries do not imply evaluation of a saved best checkpoint on an independent test set. The checkpoint callback configured to select the best validation model is not passed to the trainer in the archived training entry point.

Training accuracy, total training loss, importance loss, and the gate ratio are logged from training steps. Their final exported values represent the last recorded mini-batch statistic, not an epoch-wide or dataset-wide average. Total training loss includes the auxiliary terms, and the gated and baseline objectives therefore differ. Validation checkpoints and training steps are not independent experimental replicates; no standard deviations, confidence intervals, or significance tests across runs can be inferred from their number.

The available records contain no prediction probabilities, example-level classifications, or independent test outputs. They do not support retrospective calculation of area under the receiver operating characteristic curve (AUROC), area under the precision--recall curve (AUPRC), sensitivity, specificity, F1 score, or confusion matrices. The reanalysis uses the exported scalar values without smoothing or creation of additional training results.

\section{Results}
\subsection{Recorded Validation Performance}
\begin{figure}[!htbp]
\centering
\includegraphics[width=\linewidth]{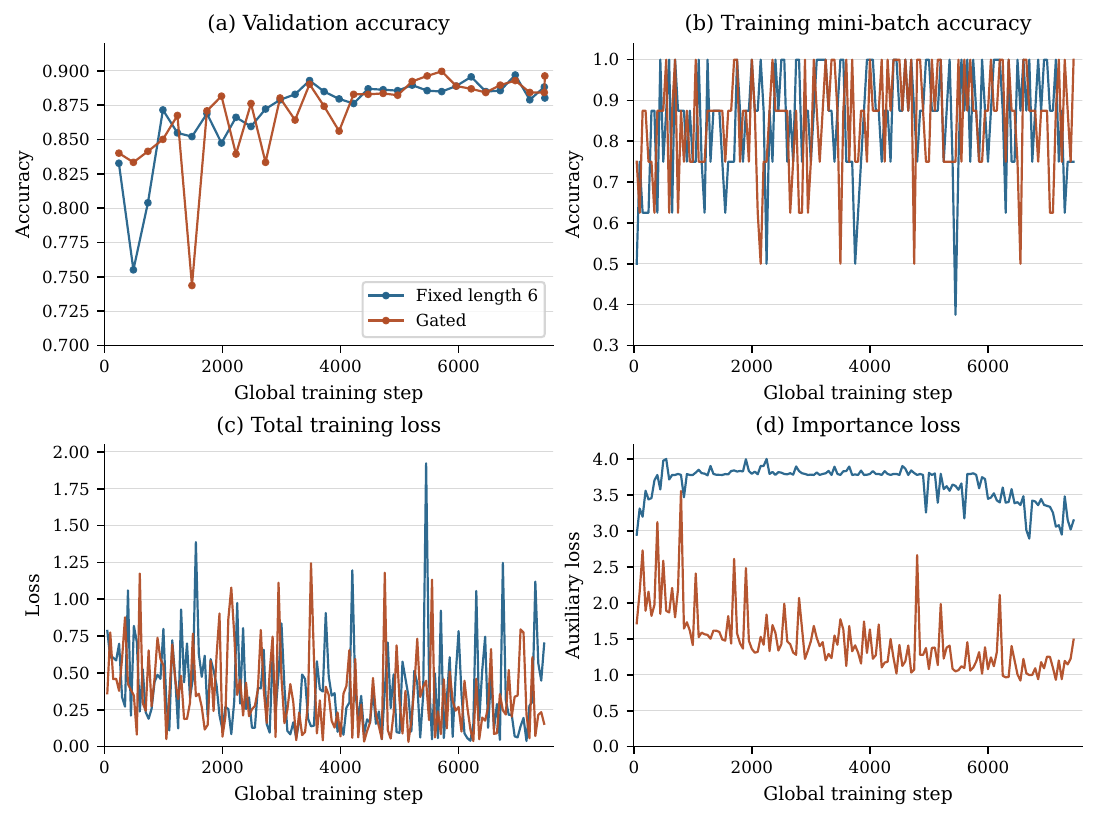}
\caption{Unsmoothed exported scalar trajectories for the fixed-length baseline and gated run. (a) Validation accuracy over all 31 recorded evaluations. (b) Training mini-batch accuracy. (c) Total training loss, including the configuration-specific auxiliary terms. (d) Importance loss. Panels (b--d) each contain 149 recorded training points; these are neither epoch averages nor repeated independent runs. The vertical axis in panel (a) is restricted to 0.70--0.92 for readability.}
\label{fig:training}
\end{figure}

The gated run reached a best validation accuracy of 0.8996, compared with 0.8969 for the fixed-length baseline (Table~\ref{tab:results}). The difference, computed from the unrounded log values, is 0.2677 percentage points. The best records occurred at different optimization steps: 5713 for the gated run and 6959 for the baseline. At the final recorded validation point, accuracy was 0.8963 and 0.8802, respectively, a difference of 1.6064 percentage points. Figure~\ref{fig:training} presents the complete exported trajectories.

The best recorded values are numerically close. This single-run comparison provides neither a superiority test nor an equivalence test. The final recorded mini-batch accuracy of 1.0000 describes one batch and cannot establish either perfect training-set performance or overfitting by itself.

\begin{table}[H]
\centering\small
\caption{Scalar summaries from one recorded run per configuration. Best and final validation values summarize the same repeatedly evaluated validation subset. Training statistics are final exported mini-batch values.}
\label{tab:results}
\begin{tabular}{lrr}
\toprule
Measure & Fixed length 6 & Gated, maximum length 6\\
\midrule
Best validation accuracy & 0.8969 & 0.8996\\
Step of best validation record & 6959 & 5713\\
Final validation accuracy & 0.8802 & 0.8963\\
Final recorded mini-batch accuracy & 0.7500 & 1.0000\\
Final recorded total training loss & 0.6998 & 0.1553\\
Final recorded importance loss & 3.1456 & 1.4875\\
\bottomrule
\end{tabular}
\end{table}

\subsection{Gate Statistics and Minimum-Length Behavior}
\begin{figure}[!htbp]
\centering
\includegraphics[width=\linewidth]{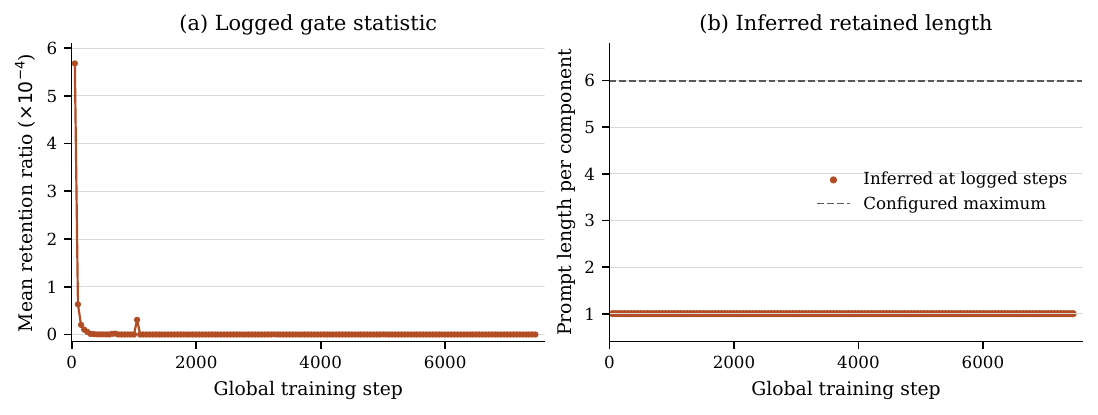}
\caption{Gate behavior at the 149 exported training points. (a) Logged mean gate ratio, averaged over the mini-batch and 12 visual layers. Values equal to zero are shown as stored, without introducing a logarithmic offset. (b) One-token retained length inferred for all visual layers at those recorded points using the non-negativity bound in Equation~\eqref{eq:bound}. The dashed line denotes the configured maximum of six. Panel (b) is an inference from the gate summary and implementation, not a direct length log or a record of unobserved batches.}
\label{fig:gate}
\end{figure}

The maximum exported gate ratio is $5.6791\times10^{-4}$, recorded at the first available training point, step 49. The final exported ratio is zero at the precision of the stored scalar. Figure~\ref{fig:gate} shows all 149 records. These values are averages over samples and the 12 visual layers, rather than directly logged per-sample lengths or per-layer length distributions.

Nevertheless, their small magnitude permits a bounded inference about the executed length at the recorded training points. If $G=H^{-1}\sum_\ell\bar r_{\mathcal{B},\ell}$ is the logged gate scalar and $G_{\max}$ is its maximum over the exported records, non-negativity gives
\begin{equation}
 0\leq\bar r_{\mathcal{B},\ell}\leq HG
 \leq12G_{\max}\approx0.006815<\frac{1}{6}.
\label{eq:bound}
\end{equation}
Substitution into Equation~\eqref{eq:length} yields $L_{\mathcal{B},\ell}=1$ for every layer at each of these logged points, provided the recorded run follows the supplied implementation. The logs do not cover the first 49 optimization steps, intervening unrecorded batches, or validation-time gate behavior. They therefore do not establish the timing of an initial transition from longer prompts, continuous one-token execution throughout training, or the evaluation-time length distribution.

\subsection{Auxiliary Loss and Resource Implications}
The final exported importance loss is 1.4875 for the gated run and 3.1456 for the baseline. This indicates a lower value of the auxiliary imbalance objective at those recorded steps; it does not establish more useful routing. The simultaneous reduction in the gate ratio also scales down the mixed prompt, so lower importance loss cannot isolate the effect of prefix length.

At an executed length of one, each mixed and static prompt component uses one-sixth of its configured token budget. The complete visual sequence, however, decreases by only ten tokens, or 4.76\% (Table~\ref{tab:budget}). The number of pairwise attention positions would decrease by $1-(200/210)^2=9.30\%$ under otherwise identical dense attention. This quantity excludes projections, feed-forward layers, the text encoder, the gate, and other execution overhead. No latency, throughput, peak-memory, or measured floating-point-operation results are available to establish a net efficiency benefit.

\section{Discussion}
\subsection{Prompt Shortening and the Meaning of Adaptivity}
The recorded experiment demonstrates a regime in which the implemented gate strongly suppresses its retention ratio while the classifier attains a validation score near that of the fixed-length baseline. It does not demonstrate sustained variation in executed prompt length. The batch-mean discretization also means that changing the other samples in a batch can change the length assigned to an otherwise unchanged sample. This batch dependence must be distinguished from an individual policy that is invariant to batch composition.

The mechanism combines several interventions: mixed-prompt truncation, static-prompt truncation, mixed-prompt amplitude scaling, and an additional gate objective. Minimum-length behavior could reflect over-allocation of the original prompt budget, strong regularization, reliance on report text, or a combination of these factors. A fixed one-token or two-token model is required to determine whether learning a gate provides any benefit beyond choosing a smaller prompt in advance. An amplitude-only gate and a truncation-only controller are additionally needed to attribute the effects of these coupled changes.

\subsection{Potential Failure Cases}
Minimum-length truncation may be unsuitable for harder multi-label tasks or categories that depend on subtle visual findings. If additional prompt positions provide useful adaptation capacity, prefix truncation removes that capacity uniformly within a batch; reducing the mixed-prompt amplitude can further weaken its contribution. Performance on the present derived binary target does not determine how these effects would behave on such tasks.

Batch-dependent length selection creates another possible failure mode. In a heterogeneous batch, an individual sample cannot retain a longer prompt independently of its neighbors. Near a discretization boundary, changing batch composition or batch size can alter the length assigned to an unchanged sample. Reliance on report features may also reduce transferability when reports are absent, incomplete, or inconsistent with the image. These are mechanism-based hypotheses for further evaluation, not observed failure rates in the available records.

\subsection{Limitations and Required Validation}
The principal limitation is the coupling of the target to report text that also enters the classifier. The sequential row partition and uncertain image-read integrity further restrict interpretation. A defensible evaluation requires a verified cohort manifest, explicit class counts, randomized and stratified study-level partitions, and patient grouping when a reliable mapping is available. Image-only, report-only, and combined-input controls are needed to assess the visual contribution. Replacing the label rule or partition defines a new experimental protocol; the present scores cannot be relabeled as results under that protocol.

The experimental evidence consists of one run per configuration with no independent test set. Repeated matched seeds, fixed short-prompt controls, and ablations of the gate-loss weight and minimum length remain necessary. Example-level probabilities and labels should support AUROC, AUPRC, sensitivity, specificity, F1 score, and confusion matrices, with model selection restricted to validation data. Direct per-sample and per-layer traces are needed to evaluate gate behavior beyond the conditional inference from the scalar summaries.

Practical efficiency remains unmeasured. The controller increases the analytical trainable-parameter count even when fewer tokens are executed. Latency, throughput, peak memory, and operation counts should be compared at matched hardware, batch size, precision, warm-up, and synchronization settings. Routing noise should be handled consistently across comparisons; disabling it would define a changed implementation requiring new measurements. Finally, transfer to a larger resource such as MIMIC-CXR requires a consistently defined target and input protocol \cite{mimic}. Cohort manifests, source revisions, checkpoints, environments, and prediction-level outputs should accompany these subsequent experiments.

\section{Conclusions}
This pilot study examines a layer-wise gate combining batch-dependent visual prompt truncation with sample-specific mixed-prompt scaling in a MoPE-based classifier. The recorded gated and fixed-length runs achieved best validation accuracies of 0.8996 and 0.8969, respectively. Gate summaries imply minimum-length execution at all recorded training points under the supplied implementation. The supported finding is prompt shortening in this weakly labeled image--text setting; an advantage over fixed short prompts, practical acceleration, and clinical performance remain unestablished. The next evaluation should compare gate-free short prompts and unimodal controls under a protocol that separates target construction and model selection from testing.

\end{document}